\documentclass[10pt,a4paper]{article}
\usepackage[utf8]{inputenc}
\usepackage[T1]{fontenc}
\usepackage{amsmath}
\usepackage{amsfonts}
\usepackage{amssymb}
\usepackage{lmodern}
\usepackage{subfiles}
\usepackage{algpseudocode}
\usepackage{algorithm}
\usepackage{tikz}
\usepackage{cleveref}
\usepackage{mathtools}
\usepackage[backend=biber]{biblatex}
\usepackage{cleveref}
\usepackage{float}
\usepackage{multirow}

\newcounter{defcounter}
\usetikzlibrary{shapes.geometric,arrows,chains,matrix,positioning,scopes,calc}
\tikzstyle{mybox} = [draw=white, rectangle]
\definecolor{darkblue}{rgb}{0,0.08,0.45}
\definecolor{blue}{rgb}{0,0,1}
\usetikzlibrary{fit,positioning}

\newcommand{\Func}[2]{#1 \left( #2\right)}
\newcommand{\FuncSq}[2]{#1 \left[ #2 \right]}
\newcommand{\CondFunc}[3]{#1 \left(#2 \, \middle\vert \, #3 \right)}
\newcommand{\Prob}[1]{\Func{p}{#1}}

\newcommand{\Cond}[2]{\CondFunc{p}{#1}{#2}}

\newcommand{\dNorm}[1]{\Func{\mathcal{N}}{#1}}

\newcommand{\dBern}[1]{\Func{\textrm{Bernoulli}}{#1}}

\newcommand{\PTheta}[1]{\Func{p_\theta}{#1}}
\newcommand{\CondPTheta}[2]{\CondFunc{p_\theta}{#1}{#2}}

\newcommand{\Expect}[2]{\FuncSq{\mathbb{E}_{#1}}{#2}}
\newcommand{\DivKL}[2]{\Func{\textrm{D}_{KL}}{#1 \, \mid \mid \, #2}}

\newcommand\myeq{\stackrel{\mathclap{\normalfont\mbox{def}}}{=}}

\usepackage{geometry} 
\usepackage[parfill]{parskip} 
\usepackage{graphicx}

\usepackage{xcolor}
\usepackage{listings}
\definecolor{codegreen}{rgb}{0,0.6,0}
\definecolor{codegray}{rgb}{0.5,0.5,0.5}
\definecolor{codepurple}{rgb}{0.58,0,0.82}
\definecolor{backcolour}{rgb}{0.95,0.95,0.92}
\usepackage{tocloft}
\makeatletter
\renewcommand{\@seccntformat}[1]{\csname the#1\endcsname.\quad}
\makeatother

\usepackage{titlesec}

\begin{document}
\begin{center}
\Large{\textsc{
Least Square Variational Bayesian Autoencoder with Regularization }}\\
\vspace{.5cm}
\large{Gautam Ramachandra}\\
\large{gautamrbharadwaj@gmail.com}\\
\large{Indian Institute of Science, Bangalore}\\
\end{center} 
\begin {abstract}
   In recent years Variation Autoencoders have become one of the most popular unsupervised learning of complicated distributions.
 Variational Autoencoder (VAE) provides more efficient reconstructive performance over a traditional autoencoder. Variational auto enocders make better approximaiton than MCMC. The VAE defines a generative process in terms of ancestral sampling through a cascade of hidden stochastic layers. Variational autoencoder is trained to maximise the variational lower bound. Here we are trying maximise the likelihood and also at the same time we are trying to make a good approximation of the data. Its basically trading of the data log-likelihood and the KL divergence from the true posterior. This paper describes the scenario in which we wish to find a point-estimate to the parameters $\theta$ of some parametric model in which we generate each observations $\mathbf{x}_i$ by first sampling a local latent variable $\mathbf{z}_i \sim \PTheta{\mathbf{z}}$ and then sampling the associated observation $\mathbf{x}_i \sim \CondPTheta{\mathbf{x}}{\mathbf{z}}$. Here we use least square loss function with regularization in the the reconstruction of the image, the least square loss function was found to give better reconstructed images and had a faster training time. \end {abstract}
\section{Introduction}
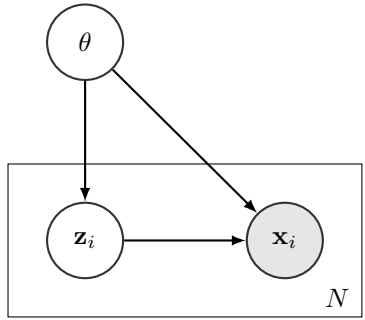
\begin{figure}[!htbp]
\centering
\begin{tikzpicture}
\tikzstyle{main}=[circle, minimum size = 10mm, thick, draw =black!80, node distance = 16mm]
\tikzstyle{connect}=[-latex, thick]
\tikzstyle{box}=[rectangle, draw=black!100]
  \node[main, fill = white!100] (theta) {$\theta$};
  \node[main, fill = white!100, below=of theta] (C1) {$\mathbf{z}_i$};
  \node[main, fill = black!10, right=of C1] (X1) {$\mathbf{x}_i$};
  \path (theta) edge [connect] (C1)
        (C1) edge [connect] (X1)
        (theta) edge [connect] (X1);
  \node[rectangle, inner sep=0mm, fit= (C1) (X1),label=below right:$N$, yshift=0mm, xshift=12mm] {};
  \node[rectangle, inner sep=5mm,draw=black!100, fit= (C1) (X1), yshift=0mm] {};
  \path
    ([shift={(50\pgflinewidth,-50\pgflinewidth)}]current bounding box.south west)
    ([shift={( 50\pgflinewidth, -150\pgflinewidth)}]current bounding box.north east);
\end{tikzpicture}
\caption{\label{fig:graph}Graphic Model Each $i^{th}$ observation is conditionally independent given the model parameters $\theta$.}
\end{figure}

 Variational Autoencoder (VAE) provides more efficient reconstructive performance over a traditional autoencoder. They are a directed graphic models. Variational autoencoder is trained to maximise the variational lower bound. Its basically trading of the data log-likelihood and the KL divergence from the true posterior. This paper describes the scenario in which we wish to find a point-estimate to the parameters $\theta$ of some parametric model in which we generate each observations $\mathbf{x}_i$ by first sampling a ``local'' latent variable $\mathbf{z}_i \sim \PTheta{\mathbf{z}}$ and then sampling the associated observation $\mathbf{x}_i \sim \CondPTheta{\mathbf{x}}{\mathbf{z}}$.  The model is graphically represented below \ref{fig:graph}.

The posterior $\CondPTheta{\mathbf{z}_i}{\mathbf{x}_i}$ is intractable for a continuous latent space whenever either the prior $\PTheta{\mathbf{z}_i}$ or the likelihood $\CondPTheta{\mathbf{x}_i}{\mathbf{z}_i}$ are non-Gaussian, meaning that approximate inference is required. To this end Autoencoding Variational Bayes makes two contributions in terms of methodology, introducing a differentiable stochastic estimator for the variational lower bound to the model evidence, and using this to learn a recognition model to provide a fast method to compute an approximate posterior distribution over ``local'' latent variables given observations.

\section{Stochastic Variational Inference}

The aim of variational inference is to provide a deterministic approximation to an intractable posterior distribution by finding parameters $\phi$ such that $\DivKL{\Func{q_\phi}{\mathbf{\theta}}}{\Cond{\mathbf{\theta}}{D}}$ is minimised. This is achieved by noting that
\begin{align}
  \DivKL{\Func{q_\phi}{\mathbf{\theta}}}{\Cond{\mathbf{\theta}}{D}} =& \log \Prob{D} + \Expect{\Func{q_\phi}{\mathbf{\theta}}}{\log \Func{q_\phi}{\mathbf{\theta}}  - \log \Prob{\mathbf{\theta} ,D}} \nonumber \\
  =:& \log \Prob{X} - \Func{\mathcal{L}}{\phi; D}.
\end{align}
Noting that $\log \Prob{D}$ is constant w.r.t. $\phi$, we can now minimise the KL-divergence by maximising the evidence lower bound (ELBO) $\mathcal{L}$ (that this is indeed a lower bound follows from the non-negativity of the KL-divergence). Aside from some notable exceptions (eg. \cite{1}) this quantity is not tractably point-wise evaluable. However, if $\Func{q_\phi}{\mathbf{\theta}}$ and $\log \Prob{\mathbf{\theta}, D}$ are point-wise evaluable, it can be approximated using Monte Carlo as
\begin{equation}
  \Func{\mathcal{L}}{\phi; D} \approx \frac{1}{L} \sum_{l=1}^{L} \log \Prob{\mathbf{\theta}_l ,D} - \log \Func{q_\phi}{\mathbf{\theta}_l}, \quad \mathbf{\theta}_l \sim \Func{q_\phi}{\mathbf{\theta}}
\end{equation}
This stochastic approximation to the ELBO is not differentiable w.r.t. $\phi$ as the distribution from which each $\mathbf{\theta}_l$ is sampled itself depends upon $\phi$, meaning that the gradient of the log likelihood cannot be exploited to perform inference. One of the primary contributions of the paper being reviewed is to provide a differentiable estimator for $\mathcal{L}$ that allows gradient information in the log likelihood $\CondFunc{p_\theta}{\mathbf{x}_i}{\mathbf{z}_i}$ to be exploited, resulting in an estimator with lower variance. In particular it notes that if there exists a tractable reparameterisation of the random variable $\tilde{\mathbf{\theta}} \sim \Func{q_\phi}{\mathbf{\theta}}$ such that
\begin{equation}
  \tilde{\mathbf{\theta}} = \Func{g_\phi}{\epsilon}, \quad \epsilon \sim \Prob{\epsilon},
\end{equation}
then we can approximate the gradient of the ELBO as
\begin{equation}
  \Func{\mathcal{L}}{\phi; D} = \Expect{\Prob{\epsilon}}{\log \Prob{\mathbf{\theta} ,X} - \Func{q_\phi}{\mathbf{\theta}}} \approx \frac{1}{L} \sum_{l=1}^{L} \log \Prob{\mathbf{\theta}_l ,X} - \log \Func{q_\phi}{\mathbf{\theta}_l} =: \Func{\tilde{\mathcal{L}}^1}{\phi; X},
\end{equation}
where $\mathbf{\theta}_l = \Func{g_\phi}{\epsilon_l}$ and $\epsilon_l \sim \Prob{\epsilon}$. Thus the dependence of the sampled parameters $\mathbf{\theta}$ on $\phi$ has been removed, yielding a differentiable estimator provided that both $q_\phi$ and $\log \Prob{\mathbf{\theta} ,D}$ are themselves differentiable. Approximate inference can now be performed by computing the gradient of $\tilde{\mathcal{L}}^1$ w.r.t. $\phi$ either by hand or using one's favourite reverse-mode automatic differentiation package (eg. Autograd \cite{6}) and performing gradient-based stochastic optimisation to maximise the elbo using, for example, AdaGrad \cite{2}.

One can also re-express the elbo in the following manner
\begin{equation}
  \Func{\mathcal{L}}{\phi; D} = \Expect{\Func{q_\phi}{\mathbf{\theta}}}{\log \Cond{D}{\mathbf{\theta}}} - \DivKL{\Func{q_\phi}{\mathbf{\theta}}}{\Prob{\mathbf{\theta}}}.
\end{equation}
This is useful as the KL-divergence between the variational approximation $\Func{q_\phi}{\mathbf{\theta}}$ and the prior over the parameters $\mathbf{\theta}$ has a tractable closed-form expression in a number of useful cases. This leads to a second estimator for the elbo:
\begin{equation}
  \Func{\tilde{\mathcal{L}}^2}{\phi; D} := \frac{1}{L} \sum_{l=1}^{L} \log \Cond{D}{\mathbf{\theta}_l} - \DivKL{\Func{q_\phi}{\mathbf{\theta}}}{\Prob{\mathbf{\theta}}}.
\end{equation}
It seems probable that this estimator will in general have lower variance than $\tilde{\mathcal{L}}^1$.

So far Stochastic Variational Inference has been discussed only in a general parametric setting. The paper's other primary contribution is to use a differentiable recognition network to learn to parameterise the posterior distribution over latent variables $z_i$ local to each observation $x_i$ in a parametric model. In particular, they assume that given some global parameters $\mathbf{\theta}$, $\mathbf{z}_i \sim \Func{p_\mathbf{\theta}}{\mathbf{z}}$ and $\mathbf{x}_i \sim \CondFunc{p_\mathbf{\theta}}{\mathbf{x}_i}{\mathbf{z}_i}$. In the general case the posterior distribution over each $z_i$ will be intractable. Furthermore, the number of latent variables $\mathbf{z}_i$ increases as the number of observations increases, meaning that under the framework discussed above we would have to optimise the variational objective with respect to each of them independently. This is potentially computationally intensive and quite wasteful as it completely disregards any information about the posterior distribution over the $z_i$ provided by the similarities between inputs locations $\mathbf{x}_{\neq i}$and corresponding posteriors $\mathbf{z}_{\neq i}$. To rectify this the recognition model $\CondFunc{q_\phi}{\mathbf{z}}{\mathbf{x}}$ is introduced.

Given the recognition model and a point estimate for $\mathbf{\theta}$, the ELBO becomes
\begin{align}
  \Func{\mathcal{L}}{\mathbf{\theta}, \phi; D} =& \,\, \Expect{\Func{q_\phi}{\mathbf{z_1}},...,\Func{q_\phi}{\mathbf{z}_N}}{\log \prod_{i=1}^{N} \Func{p_\mathbf{\theta}}{\mathbf{x}_i, \mathbf{z}_i} - \log \prod_{i=1}^{N} \Func{q_\phi}{\mathbf{z}}_i} \nonumber \\
  =& \sum_{i=1}^{N} \Expect{\Func{q_\phi}{\mathbf{z}_i}}{\log \Func{p_\mathbf{\theta}}{\mathbf{x}_i, \mathbf{z}_i} - \log \Func{q_\phi}{\mathbf{z}_i}} \nonumber \\
\end{align}

For this ELBO we can derive a similar result to $\tilde{\mathcal{L}}^1$, where we do not assume a closed-form solution for the KL divergence between distributions and include mini-batching to obtain an estimator for the ELBO for a mini-batch of observations
\begin{equation}
  \Func{\tilde{\mathcal{L}}^A}{\mathbf{\theta}, \phi; D} \approx \frac{N}{LM} \sum_{i=1}^{M} \sum_{l=1}^{L} \log \Func{p_\mathbf{\theta}}{\mathbf{x}_i, \mathbf{z}_{i,l}} - \log \Func{q_\phi}{\mathbf{z}_{i,l}}, \quad \mathbf{z}_{i,l} = \Func{g_\phi}{\mathbf{x}_i, \mathbf{\epsilon}_{i,l}},\,\, \mathbf{\epsilon}_{i,l} \sim \Func{p}{\mathbf{\epsilon}}
\end{equation}
where the $M$ observations in the mini-batch are drawn uniformly from the data set comprised of $N$ observations and for each observation we draw $L$ samples from the approximate posterior $\CondFunc{q_\phi}{\mathbf{z}_i}{\mathbf{x}_i}$.
Similarly, if $\CondFunc{q_\phi}{\mathbf{z}}{\mathbf{x}}$ and $\Func{p_\mathbf{\theta}}{\mathbf{z}}$ are such that the KL-divergence between them has a tractable closed-form solution then we can use an approximate bound which we could reasonably expect to have a lower variance:
\begin{align}
  \Func{\mathcal{L}}{\mathbf{\theta}, \phi; D} =& \,\,\Expect{\Func{q_\phi}{\mathbf{z}_1},...,\Func{q_\phi}{\mathbf{z}_N}}{\log \prod_{i=1}^{N} \CondFunc{p_\mathbf{\theta}}{\mathbf{x}_i}{\mathbf{z}_i}} - \DivKL{\prod_{i=1}^{N} \CondFunc{q_\phi}{\mathbf{z}_i}{\mathbf{x}_i}}{\prod_{i=1}^{N} \Func{p_\mathbf{\theta}}{\mathbf{z}_i}} \nonumber \\
  =& \sum_{i=1}^{N} \Expect{\CondFunc{q_\phi}{\mathbf{z}_i}{\mathbf{x}_i}}{\log \CondFunc{p_\mathbf{\theta}}{\mathbf{x}_i}{\mathbf{z}_i}} - \DivKL{\CondFunc{q_\phi}{\mathbf{z}_i}{\mathbf{x}_i}}{\Func{p_\mathbf{\theta}}{\mathbf{z}_i}} \nonumber \\
  \approx& \frac{N}{M} \sum_{i=1}^{M} \left[ \frac{1}{L} \sum_{l=1}^{L} \log \CondFunc{p_\mathbf{\theta}}{\mathbf{x}_i}{\mathbf{z}_{i,l}} - \DivKL{\CondFunc{q_{\phi}}{\mathbf{z}_i}{\mathbf{x}_i}}{\Func{p_\mathbf{\theta}}{\mathbf{z}_i}} \right] =: \Func{\tilde{\mathcal{L}}^B}{\mathbf{\theta}, \phi; D}, \label{eqn:Lb}
\end{align}
where $z_{i,l} = \Func{g_\phi}{\mathbf{x}_i, \mathbf{\epsilon}_{i,l}}$ and $\mathbf{\epsilon}_{i,l} \sim \Func{p}{\mathbf{\epsilon}}$.

\section{The Variational Autoencoder}

Variational Autoencoders is a generative model that belongs to the field of representation learning in Artificial Intelligence where an input is mapped into hidden representations. They combine the deep learning techniques with the bayesian theory. Variational Autoencoders often learns the joint distribution of a data by imposing some regularization and learns the latent variables that are very similar to the observed datapoints and prior distribution over latent space.
The variational autoencoders uses some of the methods described previously to optimize the objective function. We define a generative model where we assume that the $i^{th}$ observation was generated by first sampling a latent variable $\mathbf{z}_i \sim \dNorm{0, I}$ and that the each observation real-valued vector $\mathbf{x}_i \sim \dNorm{\Func{\mu_\theta}{\mathbf{z}_i},\Func{\sigma^2_\theta}{\mathbf{z}_i}}$ where
\begin{align}
  \Func{\mathbf{\mu}_\theta}{\mathbf{z}_i} =& \mathbf{h}_i W_\mathbf{\mu}^{(dec)} + \mathbf{b}_\mathbf{\mu}^{(dec)}, \\
  \log \Func{\sigma^2_\theta}{\mathbf{z}_i} =& \mathbf{h}_i W_\sigma^{(dec)} + \mathbf{b}_\sigma^{(dec)},
\end{align}
$\mathbf{h}_i \in \mathbb{R}^{1 \times D_h}$ is the output at the final hidden layer of the ``decoder MLP'', $W_\mu^{(dec)}, W_\sigma^{(dec)} \in \mathbb{R}^{D_h \times D_z}$ are matrices mapping from the $D_h$ hidden units to the $D_z$ dimensional latent space. Similarly, $\mathbf{b}_\mu^{(dec)}, \mathbf{b}_\sigma^{(dec)} \in \mathbb{R}^{1 \times D_z}$ are row vector biases. Note that the variances are parameterised implicitly through their logs to ensure that they are correctly valued only on the positive reals.

Here $x_i \sim \dBern{\Func{f_\theta}{\mathbf{z}_i}}$ and the output at the final hidden layer of the decoder is
\begin{equation}
  \Func{f_\theta}{\mathbf{z}_i} = \left[1 + \Func{\exp}{- \mathbf{h}_i W^{(dec)} - \mathbf{b}^{(dec)}} \right]^{-1},
\end{equation}
where $W^{(dec)} \in \mathbb{R}^{D_h \times D_z}$ and $b^{(dec)} \in \mathbb{R}^{1 \times D_z}$.

The model is given by
\begin{equation}
  \CondFunc{q_\phi}{\mathbf{z}}{\mathbf{x}} = \mathcal{N}(\mathbf{z}; \mu_\phi, \mathbf{x}_i, \sigma_{\phi^2}(\mathbf{x}_i)),
\end{equation}
where the parameters $\Func{\mu_\phi}{\mathbf{x}_i}$ and $\Func{\sigma_\phi^2}{\mathbf{x}_i}$ are again given by an MLP, which will be referred to as the encoders, whose input is $\mathbf{x}_i$.

So there is a nearby solution for the KL-divergence between the variational posterior $\CondFunc{q_\phi}{\mathbf{z}}{\mathbf{x}}$ and the priori $\Func{p_\theta}{\mathbf{z}}$. Hence we can equate the lower bound using equation 9 which is 

\begin{equation}
  \Func{g_\phi}{\mathbf{x}, \epsilon} = \Func{\mu_\phi}{\mathbf{x}_i} + \Func{\sigma_\phi}{\mathbf{x}_i} \odot \epsilon, \quad \epsilon \sim \dNorm{0, I}.
\end{equation}

\begin{algorithm}[!bthp]
\caption{Procedure by which inference is performed in the Variational Autoencoder.}
\label{alg:vae}
\begin{algorithmic}
\State $\theta, \phi \gets \textrm{Initialisation}.$
\While {$\textrm{not converged in } \theta, \phi$}
  \State {Pick subset of size $\mathbf{x}_{1:M}$ from the dataset at random.}
  \State {Compute $\mu_{1:M}, \log \sigma_{1:M}^2$ }
  \State {For all $i \in \{1,...,M\}$, sample $\epsilon_{i} \sim\dNorm{0, I}$.}
  \State {For all $i \in \{1,...,M\}$, $\mathbf{z}_i \gets \mu_i + \sigma_i \odot \epsilon_i$.}
  \State {For all $i \in \{1,...,M\}$, compute $\log \Cond{\mathbf{x}_i}{\mathbf{z}_i}$ using the decoder and likelihood function.}
  \State {Compute $g \gets \nabla_{\theta, \phi} \Func{\tilde{\mathcal{L}}^B}{\theta, \phi; \mathbf{x}_{1:M}, \epsilon_{1:M}}$}
  \State {Update $\theta, \phi$ using gradient estimate $g$.}
\EndWhile
\end{algorithmic}
\end{algorithm}

\section{Replications and Extensions}

\subsection{Visualization}
\begin{figure}[!htb]
\minipage{0.5\linewidth}%
\begin{center}
\includegraphics[width=0.6\linewidth]{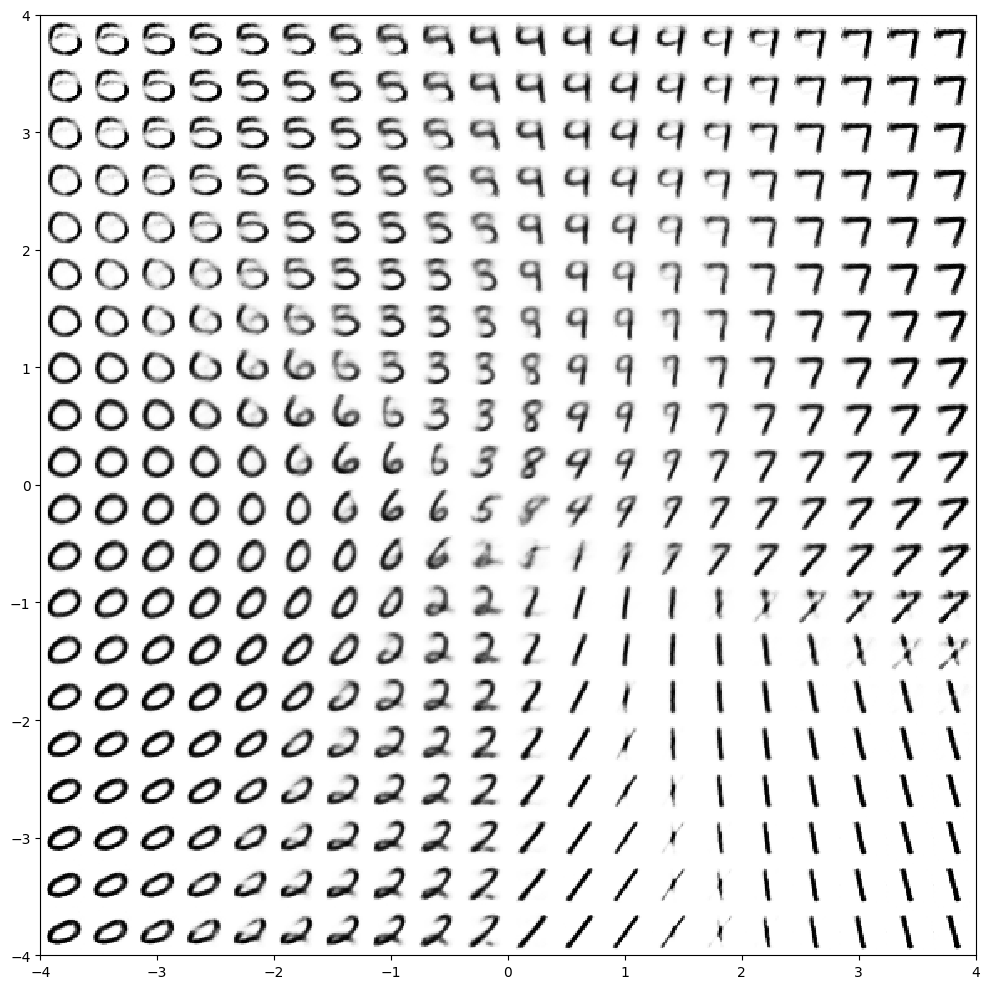}
\end{center}
\endminipage 
\minipage{0.5\linewidth}  
\begin{center}
\includegraphics[width=0.5\linewidth]{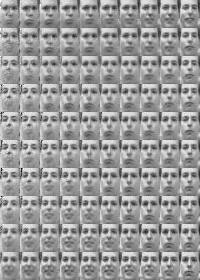}
\end{center}
\endminipage\hfill

  \caption[1]{Visualisations of learned data manifolds using two-dimensional latent space, trained with AEVB and Gaussian prior over the latent variable $\mathbf{z}$.}
\label{fig:FREY}
\end{figure}

Here we show the the learnt representations of encoders by mapping the latent space in $2D$,$3D$,$4D$ and $5D$. Here we have mapped the coordinates through inverse CDF  $\mathbf{z}$. Then, we plotted the output of our decoder $p_{\mathbf{\theta}} (\mathbf{x}| \mathbf{z})$ with the estimated parameters $\mathbf{\theta}$. Figure \ref{fig:FREY} shows the results for two learned manifolds.

In the case of the MNIST dataset, the visualisation of learned data mainfolds can be seem. In the case of the second dataset, it is interesting that the first half of the manifold shows the face from left profile while the second half slowly transforms it to the right profile. 

\subsection{Numeber of Samples}
\begin{figure}[!htb]
\minipage{0.5\linewidth}%
\begin{center}
\includegraphics[width=0.7\linewidth]{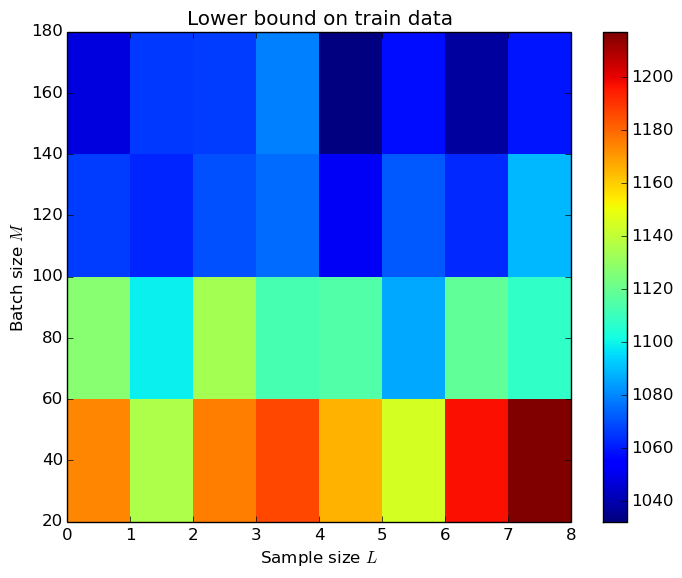}
\end{center}
\endminipage 
\minipage{0.5\linewidth}  
\begin{center}
\includegraphics[width=0.7\linewidth]{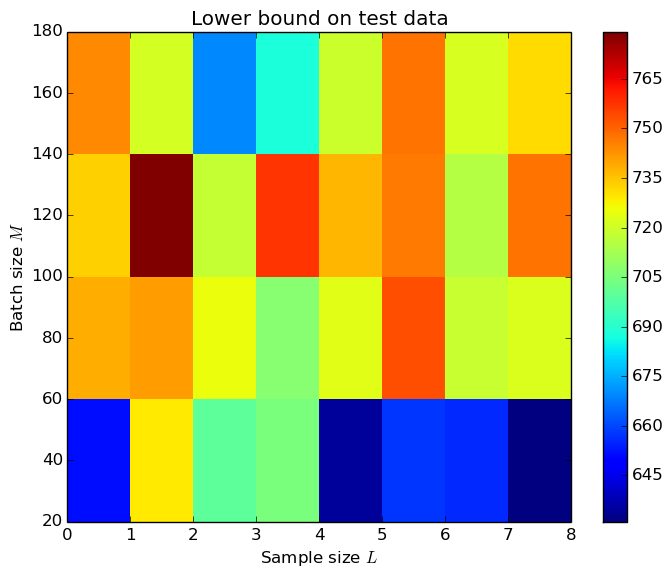}
\end{center}
\endminipage\hfill
  \caption[1]{Heatmaps of lower bound for Frey Face data set.}
  \label{fig:heatmaps}
\end{figure}

It was found out that the number of samples $L$ per data point can be set to $1$ as long as the mini-batch size $M$ was large enough, e.g. $M = 100$. However, they didn't provide any empirical results and we decided to run the comparison between the number of samples and a batch size in terms of optimizing the lower bound. Due to the computational requirements ($32$ models needed for evaluation), the experiment was run only with Frey Face dataset with $2000$ epochs of training. Figure \ref{fig:heatmaps} presents the results for sample size ranging from $1$ to $8$ and batches of size $20$, $60$, $100$ and $140$.

In the case of the test data, the highest score was obtained with $L=2$ and $M=100$ and the score was invariant to the sample size only with the batch size larger than $20$. In the case of training set, the highest score for lower bound was obtained with the batch of size $20$ where the sample size $L$ makes a big influence. For larger batch sizes, the sample size becomes invariant in terms of the score for the lower bound. However, it is possible that the models trained with a larger batch size might need more time to converge. So we used least square loss function with regularization that provided better convergence.

\subsection{Increasing the depth of the encoder}
Extending the depth of the neural networks proved to be a very helpful method in increasing the power of neural architectures. We decided to add additional hidden layers to test how much gains can be obtained in terms of optimizing the lower bound and at the same time still obtaining robust results. The experiment was run with MNIST data set having $500$ hidden units with the size of the latent space $N_z$ set to $10$ which seems to be optimal value. 

\begin{figure}[!htb]
\centering
\includegraphics[width=0.8\linewidth]{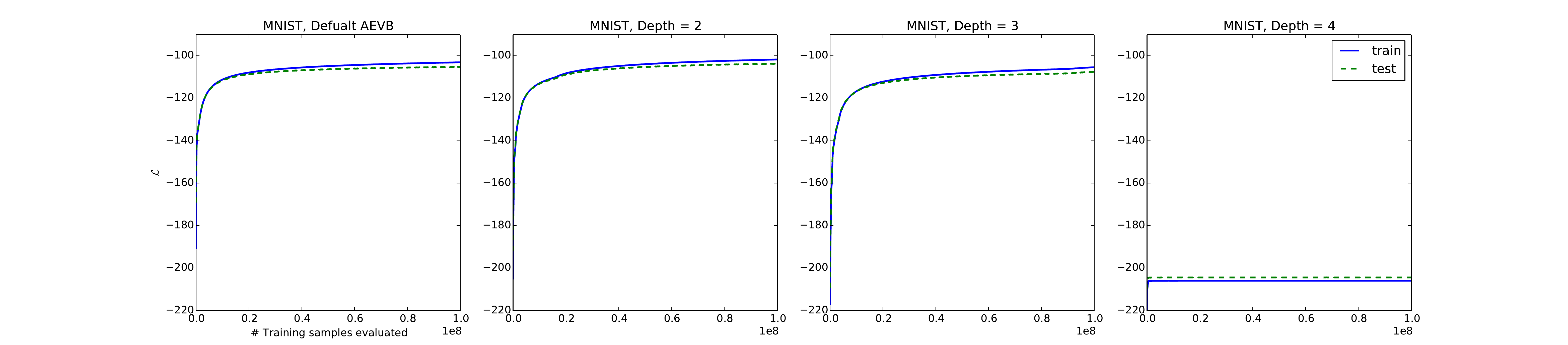}
  \caption[1]{Comparison of performance between different encoders architectures in terms of optimizing the lower bound with dimensionality of latent a space set to $10$. }
\end{figure}

Additional hidden layers didn't yield substantial increase in the performance however the encoder with two hidden layers performed slightly better than the original architecture. Presumably owing to the ``vanishing gradients'' problem, adding fourth hidden layer resulted in the inability of the network to learn.

\subsection{Noisy KL-divergence estimate}
Until now we were assuming that the $KL$-divergence term $D_{KL} (q_{\phi}(\mathbf{z} | \mathbf{x}^{(i)}) || p_{\boldsymbol{\theta}}( \mathbf{z}) ) $ can be obtained using a closed-form expression. However, in the case of the non-Gaussian distributions it is often impossible and this term also requires estimation by sampling. This more generic SGVB estimator $\widetilde{\mathcal{L}}^{A}(\boldsymbol{\theta}, \boldsymbol{\phi}; \mathbf{x}^{(i)})$ is of the form:

$$ \widetilde{\mathcal{L}}^{A}(\boldsymbol{\theta}, \boldsymbol{\phi}; \mathbf{x}_{i}) = \frac{1}{L} \sum_{l=1}^L \left( \log p_{\boldsymbol{\theta}}(\mathbf{x}_{i}, \mathbf{z}_{i,l}) - \log q_{\phi}(\mathbf{z}_{i,l} | \mathbf{x}_i) \right).$$

\begin{figure}[!htb]
\centering
\includegraphics[width=0.8\linewidth]{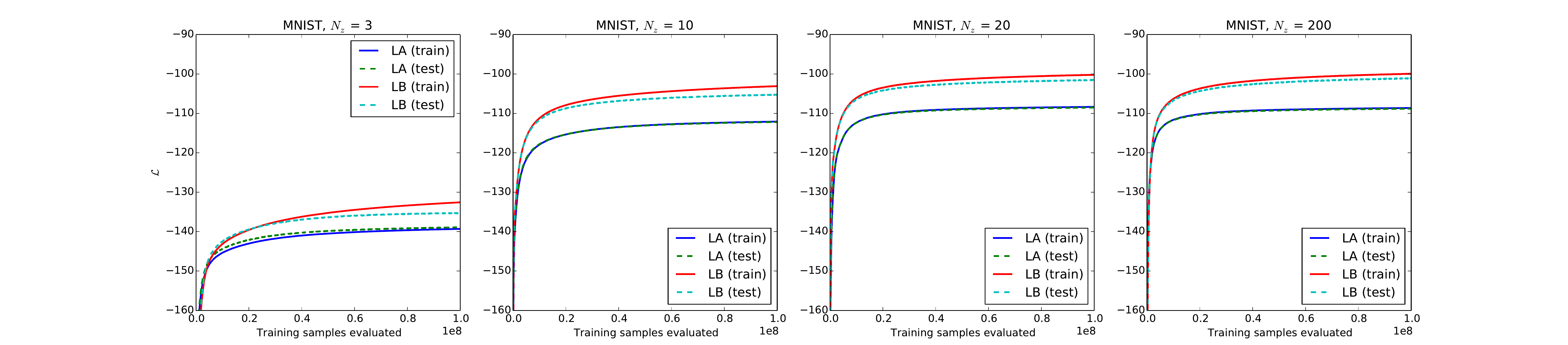}

  \caption[1]{Comparison of two SGVB estimators in terms of optimizing the lower bound for different dimensionality of latent space ($N_z$). }
  \label{fig:mnist_LAvsLB}
\end{figure}

Naturally, this form in general will have greater variance than the previous estimator. We decided that it will be informative to compare the performance of both estimators using only one sample i.e. $L=1$ -- this will allow us to observe how much more robust is the first estimator. Figure \ref{fig:mnist_LAvsLB} shows the results for the MNIST data set.

As we can see, in each case the performance of the $\widetilde{\mathcal{L}}^{B}$ estimator is substantially better. Moreover, this shows that the generic estimator might be used if we increase the sample size $L$ to reduce the variance of the estimator, however, this needs to be examined empirically. Additionally, this comes with higher computational costs.

Although, as it was expected, ReLu function learns the fastest, there is no substantial gains over tanh function. In each case, the sigmoid function learns the slowest and obtains the lowest score for the bound and at the same time the training time took about $20\%$ more time than in the case of the two other functions.

\section{Reconstruction}

The VAE with L2 regularization (VAE) provides a probabilistic approach to modeling the data as a distribution around some underlying manifold.  This allows us to define a posterior distribution over the latent space, and a distribution over input space which can be used to score inputs, while being able to measure uncertainty over both estimates. Here we choose to represent traditional Variational Autoenocder as AE and Variational Autoencoders with L2 Regularisation as VAE. The Variational autoencoders (AE) takes an input vector and maps it to some latent space, however only providing a point estimate on some lower dimensional manifold. VAEs are trained by minimizing the reconstruction error of training examples usually measured by mean squared error (MSE), and a weight regularizer:
\begin{equation}
E(\mathbf{x, \theta}) = \lVert \mathbf{x} + \sigma_\theta(\mathbf{x})\rVert^2 + \lambda \ell(\mathbf{\theta}),
\end{equation}
where $\sigma(\cdot)$ represents the application of the autoencoder (encoding followed by decoding) to a particular data example, $\ell(\cdot)$ represents a specific weight decay, and $\lambda$ represents the weight penalty. This can be seen to resemble the variational lower bound:
\begin{equation}
\Func{\mathcal{L}}{\mathbf{\theta}, \phi; \mathbf{x}_i}=\Expect{\Func{q_\phi}{\mathbf{z}_i|\mathbf{x}_i}}{\log \CondFunc{p_\mathbf{\theta}}{\mathbf{x}_i}{\mathbf{z}_i}} - \DivKL{\CondFunc{q_\phi}{\mathbf{z}_i}{\mathbf{x}_i}}{\Func{p_\mathbf{\theta}}{\mathbf{z}_i}},
\end{equation}
in which the first term can be seen as the expected negative reconstruction weight and the second  acts as a regularizer pulling latent variables towards the prior. Rather than defining a distribution in latent variable space, the AE instead provides a point estimate in the bottleneck layer, but the two are analogous in that they provide a lower dimensional representation of an input.\\

Next we add L2 regularisation as loss function for the Variational Autoencoders. Since the latent representation in the VAE with L2 regularisation is a distribution we have to choose how we use this to create a single decoded example. Both using the mean of the distribution and averaging the output of multiple samples were tried. Using the mean was shown to give an absolute improvement in error of about 0.2\%, and so that is the method that is used for further experiments.\\

Instantiations of each model type were trained using the configurations: the encoder and decoder each have one hidden layer of size 500 consisting of, tanh activation functions and using the same $\ell_2$ normalization penalty on the same portion of the MNIST data set. The resulting mean construction errors for the two models for a different size of the latent space/ compression layer are shown in Figure \ref{fig:AE_MSE}.

\begin{figure}[hbt]
\centering
\hspace{0em}
\includegraphics[width=0.4\columnwidth]{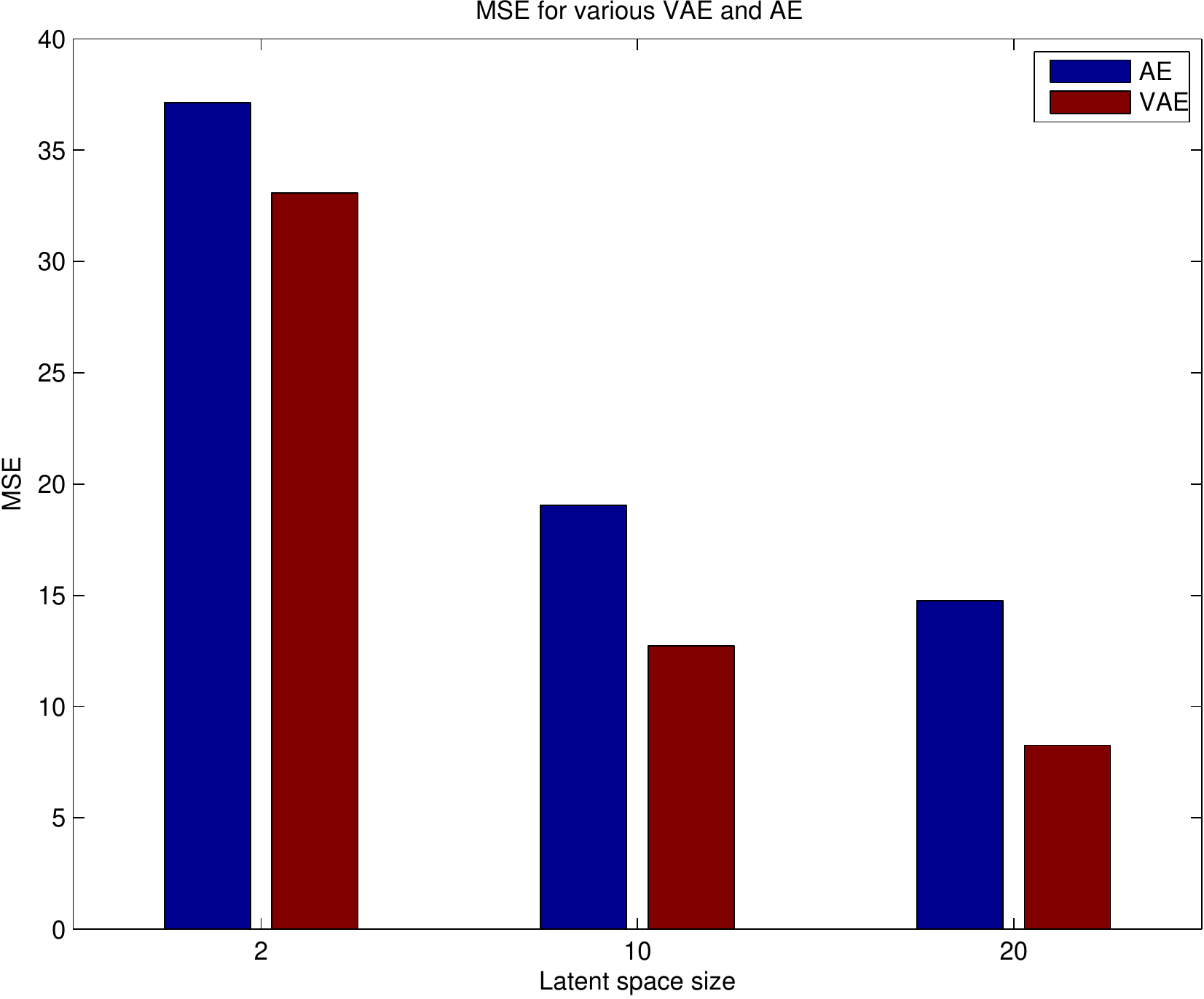}
\caption{Comparison of reconstruction error measured by MSE for variational auto-encoder (AE) and variational auto-encoder with L2 regularization (VAE) for various sizes of representation space}
\label{fig:AE_MSE}
\end{figure}

\begin{figure}[hbt]
\centering
\hspace{0em}
\includegraphics[width=0.4\columnwidth]{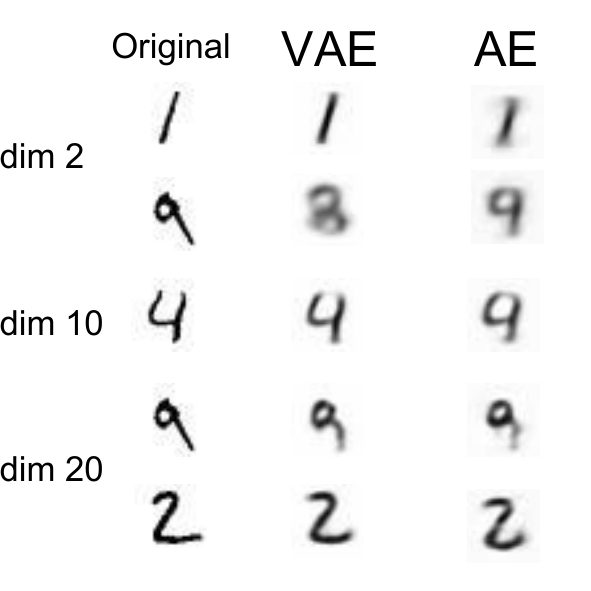}
\caption{Examples of the quality of reconstruction of certain digits for VAE with L2 regularization and VAE}
\label{fig:AE_recon}
\end{figure}

It can be seen that the variational auto-encoder with L2 regularization outperforms the normal variational auto-encoder for all reduced dimensional sizes. The difference in reconstruction error increases as the size of the latent space/ bottleneck layer increases. This suggests that the difference in performance is due to the better generalisation afforded by the variational bayesian approach.\\ Note that the VAE is directly trained to minimize the criterion that we have used to compare the two models, whereas VAE with L2 regularization is trained to minimize the expected reconstruction rate, which for the discrete case is the binary cross entropy. So despited having an ``unfair advantage'' the auto-encoder still performs worse.\\
To further contrast the two approaches (Figure \ref{fig:AE_recon}) shows specific examples of digits constructed by VAE and VAE with L2 regularization. It can be seen that generally the VAE produces images that are sharper and more similar to the original digit, with the exception of the number 9 for the 2 dimensional case. It was initially speculated that this would correspond to a higher variance of the posterior, however this was found to not be the case. Looking at the representation of the two dimensional latent space in $?$ we can see that the further right we go the more rightward slanting nines we have, so having a leftward slanting nine would push us away from nines towards the region of weight space containing eights. In such a compressed latent space it seems reasonable to assume that there will be forms of certain digits that are unrepresentable, in this case we have found an unfortunate example on which the VAE performs poorly.\\

The reconstructed images of VAE with L2 regularisation with more training epochs is shown below. It can be see that the first row has sharper image quality than the second.

\begin{figure}[hbt]
\centering
\hspace{0em}
\includegraphics[width=0.4\columnwidth]{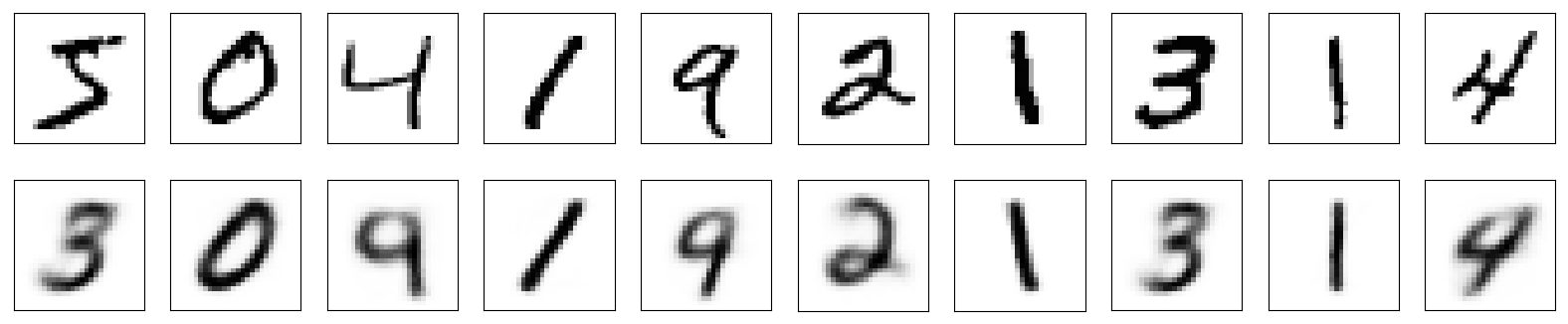}
\caption{First row is the reconstructed image of VAE with L2 regularisation and second is the normal VAE }
\label{fig:AE_MSE}
\end{figure}

\begin{figure}[hbt]
\centering
\hspace{0em}
\includegraphics[width=0.4\columnwidth]{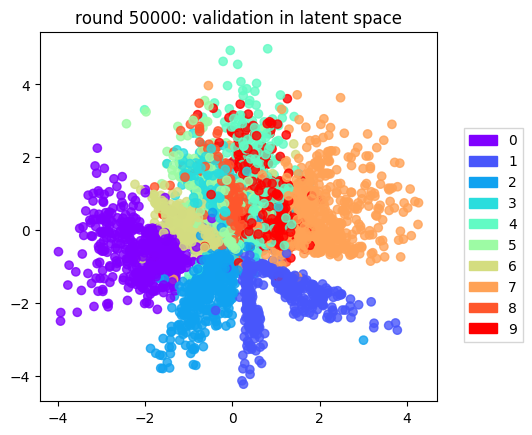}
\caption{ This is how the encoder/inference network learns to map the training set from the input data space to the latent space. }
\label{fig:AE_MSE}
\end{figure}

\section{Full Variational Bayes}
As well as providing a method of variational inference over the parameters of a latent space Kingma and Welling also detail a method of performing full variational Bayesian inference over the parameters. In this scheme we place a hyperprior over the parameters of the model $p_\alpha(\mathbf{\theta})$. The variational of the lower bound of the marginal likelihood can then be written:
\begin{equation}
\Func{\mathcal{L}}{\mathbf{\phi}; \mathbf{X}}=\Expect{\Func{q_\phi}{\theta}}{\log p_\mathbf{\theta}(\mathbf{X})} - \DivKL{q_\phi(\mathbf{\theta})}{\Func{p_\mathbf{\alpha}}{\mathbf{\theta}}}
\label{eq:full_vb_ml}
\end{equation}
By maximizing this we are encouraging the model to reconstruct the data accurately, while constraining the form that the distribution of parameters can take. For a particular point we have a variational lower bound on the marginal likelihood:
\begin{equation}
\Func{\mathcal{L}}{\mathbf{\theta}, \mathbf{\phi}; \mathbf{x^{(i)}}} =\Expect{\Func{q_\phi}{\mathbf{z} | \mathbf{x^{(i)}}}}{\log p_\mathbf{\theta}(\mathbf{z} | \mathbf{x^{(i)}})} - \DivKL{q_\phi(\mathbf{z} | \mathbf{x^{(i)}})}{\Func{p_\mathbf{\theta}}{\mathbf{z}}}
\label{eq:full_vb__dp_ml}
\end{equation}
Combining \cref{eq:full_vb_ml} and \cref{eq:full_vb__dp_ml}, using the same reparameterization trick as $\tilde{z}=g_{\phi}(\mathbf{\epsilon})$ with $\epsilon \sim p(\epsilon)$ and using the same trick for the variational approximation to the posterior over parameters: $\tilde{\mathbf{\theta}} = h_\phi(\zeta)$ with $\zeta = p(\zeta)$ we arrive at the differentiable Monte Carlo estimate for the variational lower bound of the marginal likelihood:
\begin{equation}
\mathcal{L}(\phi; \mathbf{X}) \approx \frac{1}{L}\sum_{l=1}^{L} N\cdot(\log p_{\tilde{\theta}}(\mathbf{x} | \mathbf{\tilde{z}})
 + \log p_{\tilde{\theta}}(\tilde{\mathbf{z}}) - 
 \log q_\phi(\mathbf{\tilde{z}} | x)) + \log p_\alpha(\mathbf{\tilde{\theta}}) - \log q_\phi(\mathbf{\tilde{\theta}})
 \label{eq:fvb_mc_approx}
\end{equation}
which can be maximized by performing SGVB as before by differentiating with respect to $\mathbf{\phi}$.

They provide a concrete example of a realisation of the above model in which we assume standard normal distributions for the priors over the variables and latent space, and have variational approximations to the posteriors of the form:

\begin{equation}
\begin{split}
q_\phi(\mathbf{\theta}) & = \mathcal{N}(\mathbf{\theta}; \mathbf{\mu}_\mathbf{\theta}, \mathbf{\sigma^2_\theta}\mathbf{I})\\
q_\phi(\mathbf{z|x}) & = \mathcal{N}(\mathbf{z}; \mathbf{\mu}_\mathbf{z}, \mathbf{\sigma^2_z}\mathbf{I})
\end{split}
\end{equation}

Those assumptions enable us to obtain closed form solutions for the KL term. This approach was implemented and tested on the MNIST and Frey Face data sets. Although the lower bound was increased, progress was extremely slow, the training lower bound increased much faster than the validation set, and evaluation of the reconstruction ability of the resulting models showed that no learning had taken place. 

The very slow progress to an eventual poor model resembled the effects of starting in a poor region of neural network parameter space,
and so the initial values of $\mu_\sigma$ were seeded with the MAP solutions from a regular VAE trained to convergence while $\sigma_\theta^2$ were all set to be $10^{-3}$, thereby hopefully encouraging the model to learn a distribution around a known good configuration of parameters. Nonetheless, this yielded identically poor results.

The purpose of performing inference over the parameters is to reduce overfitting and promote generalization. However in the scheme proposed it appears that the model underfits to the extent that it simply does not learn. There are a number of possible explanations for this. One problem that was faced in the implementation was negative values of variances. This was worked around by using a standard deviation which is then squared to yield a positive variance. In their recent paper on variational inference over MLP parameters \cite{7} work around this by parameterizing $\sigma$ as $\sigma = \log(1 + \exp(\rho))$. Despite yielding a closed form solution, a standard normal prior over weights is perhaps too wide for MLP weight parameters, which typically have very low variance about zero. \cite{7} found that despite not yielding a closed form solution a complicated spike-and-slab-like prior performed best composed of a mixture of a high variance and low variance Gaussian centered at 0 performed well.

Performing full variational inference will allow robust weight estimates even in low resource environments. The approach in the paper favours a neat analytical form of prior over analytically complicated priors that may induce more reliable weight estimates. The trade off between precision of gradient estimates and efficacy of form is an interesting problem that requires further research.

\section{Future Works}

An obvious extension to the paper to investigate is to simply change the form of the prior and the variational approximation in an attempt to induce a particular form of latent space. For example a particularly interesting set up would be to define a sparsity inducing prior that encourages each dimension of the latent space to be approximately valued on $\{0, 1\}$. An obvious choice would be a set of sparse Beta distributions (ie. ones in which the shape parameters $\alpha, \beta < 1$), but one could also use pairs of univariate Gaussians with means $0$ and $1$ and small variances.

Such a prior would be useful for two reasons - firstly it would allow one to provide a binary encoding for a data set by truncating the posterior approximation for any particular observation to be exactly vector binary valued allowing for a large amount of lossy compression. The posterior distribution over the parameters $\theta$ and latent values $\mathbf{z}_i$ also contains rotational symmetry which may affect the quality of the approximate inference if it attempts to place posterior mass over the entirety of this. Were a prior such as the one proposed used, this rotational symmetry would be destroyed and replaced with a ``permutation symmetry'', similar to that found in a finite mixture model.
\\

We currently assume a simple parametric form for the approximate posterior $q_\phi(\mathbf{z}|\mathbf{x})$ that allows the use of the reparameterization trick. Although this yields a robust training regime, it limits the expressibility of the model to a subset of potential distributions. If instead we directly use the $g_\phi(\mathbf{x}, \epsilon)$ we can induce an arbitrarily complex posterior that would allow us to approximate any true posterior.\\
This idea has been recently realised using Gaussian processes by \cite{7} who draw random latent input samples, push them through a non-linear mapping and then draw posterior samples. If we instead were to use a MLP to model $g_\phi(\mathbf{x}, \epsilon)$ we can, theoretically, model arbitrary posteriors. The problem now is the ability to yield a differentiable distribution over latent space which can potentially be sampling multiple $g_\phi(\mathbf{x}, \epsilon_i)$ to approximate a distribution, and batching gradients over all samples. This is akin to a Monte Carlo estimate of the variational posterior.

One of the most popular approaches in the unsupervised learning using autoencoding structures is making the learned representation robust to partial corruption of the input pattern \cite{9}. This also proved to be an effective step in pre-training of deep neural architectures. Moreover, this method can be extended where the network is trained with a schedule of gradually decreasing noise levels \cite{10}.  This approach is motivated by a desire to encourage the network to learn a more diverse set of features from a coarse-grained to fine-grained ones.

Moreover, there was recently an effort to inject noise into both an input and in the stochastic hidden layer (denoising variational autoencoder, DVAE) which yields better score in terms of optimising the log likelihood \cite{11}. In order to estimate the variational lower bound the corrupted input $\mathbf{\widetilde{x}}$,  obtained from a known corruption distribution $p (\mathbf{\widetilde{x}} | \mathbf{x})$ around $\mathbf{x}$, requires to be integrated out which is intractable in the case of $\text{E}_{p (\mathbf{\widetilde{x}} | \mathbf{x})} \left[ q_{\phi}(\mathbf{z} | \mathbf{\widetilde{x}} )\right]$. Thus, Im et al. arrived at the new form of the objective function -- the denoising variational lower bound:

$$\mathcal{L}^{C} ~ \myeq ~ \text{E}_{\tilde{q}_\phi (\mathbf{z} | \mathbf{x})} \left[ 
\log \frac{p_{\theta} (\mathbf{x, z})}{q_\phi (\mathbf{z} | \mathbf{\tilde{x}})}
\right], $$

where $\tilde{q}_\phi (\mathbf{z} | \mathbf{x}) = \int q_\phi (\mathbf{z}| \mathbf{\tilde{x}}) p(\mathbf{\tilde{x}} | \mathbf{x}) \text{d}\mathbf{\tilde{x}}$. 

However, the noise in this case was set to a constant during the training procedure. To the best of our knowledge no one analysed how the scheduling scheme might influence the learning of the auto-encoder's structure as well as the approximate form of the posterior of the latent variable. We believe that combination of both scheduled denoising training with the variational form of an auto-encoder should lead to gains in terms of the optimising lower bound and improving the reconstruction error as it was the case in the section $5$.

\section{Conclusions}

In this paper we provide a clear introduction to a new methodology for performing the reconstruction of images with variational inference using L2 regularisation. We managed to obtain better results with L2 regularization for both data sets. We found the model structure very robust to changes of parameters of the network. Moreover, our experiments show that the performance of VAEB with L2 regularization is superior to that of the traditional VAE architecture and it is resistant to superfluous latent variables thanks to automatic regularisation via the KL-term. Our implementation of variational inference on both the parameters $\theta$ and the latent variables $\mathbf{z}$ performed disappointingly poorly which might be partially explained by overly-restrictive prior and we plan to further investigate this problem.

\end{document}